%% file: 00_main.tex
\documentclass[letterpaper, 10 pt, conference]{ieeeconf}  

\IEEEoverridecommandlockouts                              

\usepackage{cite}
\usepackage{url}
\usepackage{amsmath}
\usepackage{array}
\usepackage{booktabs} 
\usepackage{multirow} 
\usepackage{makecell}
\usepackage{threeparttable}
\usepackage{graphicx}
\usepackage{amsfonts}
\usepackage{pifont}
\usepackage{siunitx}
\usepackage[dvipsnames]{xcolor}
\usepackage{microtype}
\usepackage{balance}
\usepackage{svg}
\usepackage{xspace}

\newcommand{\scene}[1]{\textls[-50]{\texttt{#1}}}

\newcommand{\cmark}{\ding{51}}%
\newcommand{\xmark}{\ding{55}}%

\newcommand{\myMethod}{FUS3DMaps\xspace} 

\usepackage{hyperref}
\title{\LARGE \bf
\myMethod: Scalable and Accurate Open-Vocabulary Semantic Mapping by 3D Fusion of Voxel- and Instance-Level Layers
}

\author{Timon Homberger, Finn Lukas Busch, Jesús Gerardo Ortega Peimbert, Quantao Yang, Olov Andersson%
\thanks{This work was partially supported by the Wallenberg AI, Autonomous Systems and Software Program (WASP) funded by the Knut and Alice Wallenberg Foundation.}
\thanks{The authors are with the Division of Robotics, Perception, and Learning, KTH Royal Institute of Technology, Sweden. Contact: \texttt{\{timonh, flbusch, olovand\}@kth.se}{\tt\small}}
}%

\begin{document}

\maketitle

\thispagestyle{empty}
\pagestyle{empty}


\begin{abstract}

Open-vocabulary semantic mapping enables robots to spatially ground previously unseen concepts without requiring predefined class sets.
Current training-free methods commonly rely on multi-view fusion of semantic embeddings into a 3D map, either at the instance-level via segmenting views and encoding image crops of segments, or by projecting image patch embeddings directly into a dense semantic map.
The latter approach sidesteps segmentation and 2D-to-3D instance association by operating on full uncropped image frames, but existing methods remain limited in scalability.
We present \myMethod, an online dual-layer semantic mapping method that jointly maintains both dense and  instance-level open-vocabulary layers within a shared voxel map. 
This design enables further voxel-level semantic fusion of the layer embeddings, combining the complementary strengths of both semantic mapping approaches.
We find that our proposed semantic cross-layer fusion approach
improves the quality of both the instance-level and dense layers, while also enabling a scalable and highly accurate instance-level map where the dense layer and cross-layer fusion are restricted to a spatial sliding window.
Experiments on established 3D semantic segmentation benchmarks as well as a selection of large-scale scenes show that \myMethod achieves accurate open-vocabulary semantic mapping at multi-story building scales.
Additional material and code will be made available: \url{https://githanonymous.github.io/FUS3DMaps/}.

\end{abstract}

\input{01_introduction}

\input{02_related_work}

\input{03_method}

\input{04_experimental_setup}

\input{05_results}

\input{06_conclusion}


\bibliographystyle{IEEEtran}
\bibliography{bibtex}

\end{document}

%% file: 01_introduction.tex
\section{Introduction}

Semantic mapping is instrumental in enabling scene understanding and autonomous decision-making for mobile systems.
Traditional approaches often rely on a predefined, environment-specific set of classes~\cite{jatavallabhula2023conceptfusionopensetmultimodal3d}.
However, advances in vision foundation models~\cite{kirillov2023segany}\cite{radford2021learningtransferablevisualmodels} enable spatial grounding of open-set semantic concepts.
Training-free open-vocabulary semantic mapping methods can be grouped into methods that generate dense semantic 3D maps, storing an embedding vector for each point or voxel, and those that generate instance-level semantic maps, grouping points or voxels into instances and storing one embedding vector per instance. Typically, dense maps have a higher capacity to store fine-grained detail, whereas instance-level maps are substantially more memory efficient and scalable, as they store fewer high-dimensional embedding vectors per mapped volume~\cite{yamazaki2023openfusionrealtimeopenvocabulary3d}\cite{werby23hovsg}.

Instance-level semantic mapping approaches commonly use segmentation and encode image crops separately for each segment~\cite{gu2023conceptgraphsopenvocabulary3dscene}\cite{werby23hovsg}.
In contrast, recent dense semantic mapping approaches use patch-level semantic embeddings of a language-aligned vision model and aggregate them in a dense semantic voxel map~\cite{alama2025rayfrontsopensetsemanticray}. 
These methods are not affected by segmentation errors,
viewpoint-dependent variations in segmentation granularity of object parts,
or subsequent association errors between new segments and existing 3D instances.
However, since they do not explicitly model object instances, depending on viewing distance and angle, the image patches may not align well with object boundaries.

\begin{figure}[t]
    \centering
    \includegraphics[width=\linewidth]{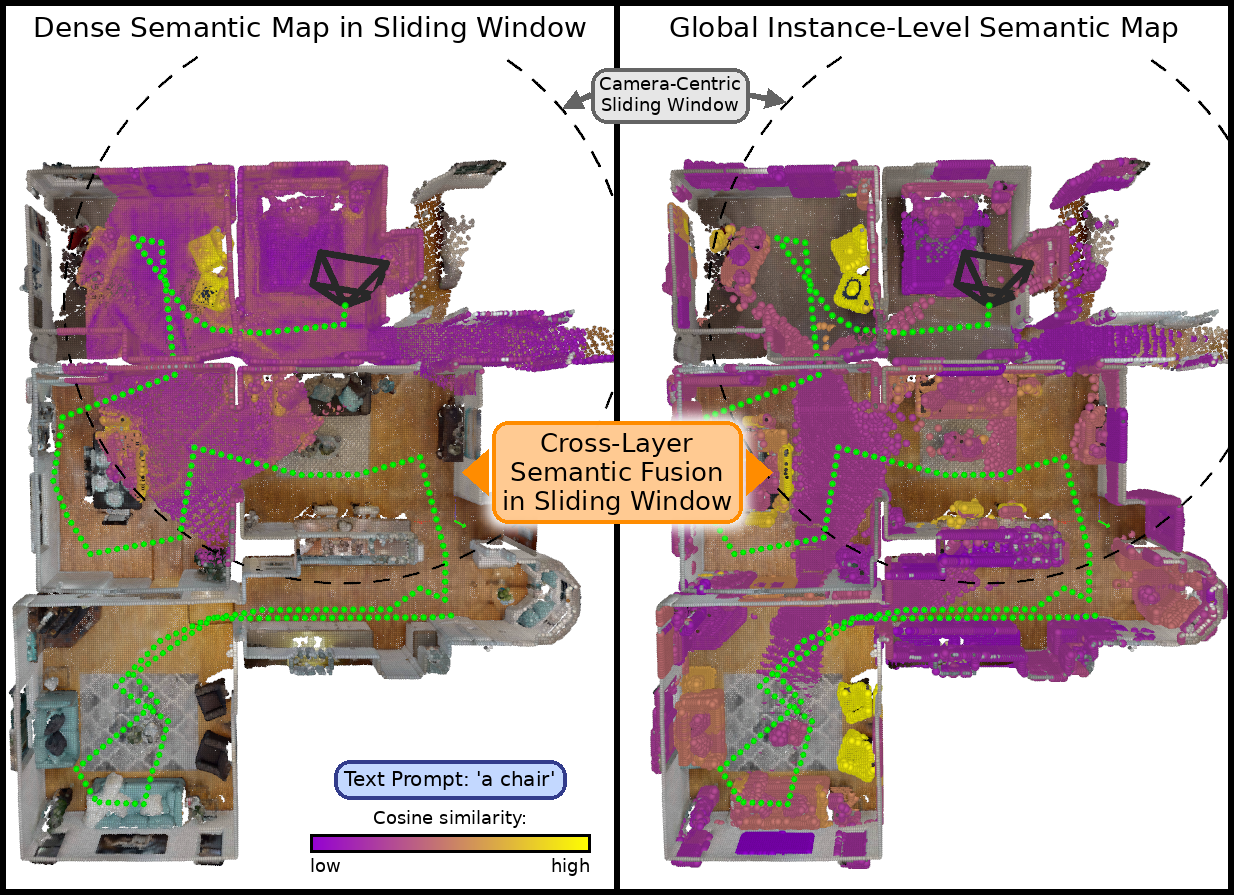}
    \vspace{-0.3cm}
    \caption{\myMethod maintains an object focused semantic map layer with instance-level embeddings. Simultaneously, it maintains a dense semantic layer, that can be restricted to a sliding window, centered at the camera pose. Semantic fusion between the dense and instance-level layers can produce mutual benefits. The sliding window approach helps to bound computational expense and reduce the memory footprint, which enables accurate semantic mapping at large scales. The visualized maps are created using scene \scene{00824} of the HM3DSem dataset~\cite{yadav2022habitat}.}
    \label{fig:hybrid}
    \vspace{-0.3cm}
\end{figure}

We propose combining the complementary strengths of instance-level and dense semantic mapping. To this end, we introduce \myMethod,
a 3D incremental open-vocabulary voxel mapping method that maintains a two-layered semantic map comprising of a dense voxel-level layer, a multi-hypothesis instance-level layer, as well as a fusion mechanism for combining embeddings of the two layers.
The proposed cross-layer semantic fusion mechanism addresses viewpoint-dependent segmentation inconsistencies by accounting for the per-voxel observation counts of each instance hypothesis.
We further propose a sliding window mechanism in which the dense layer and cross-layer fusion are restricted to a spatial window, centered at the camera pose, while the instance-level map layer is maintained globally.
This design enables enriching the instance-level embeddings with embeddings from the dense semantic layer while retaining the favorable scalability properties of instance-level maps.
We show that our fusion approach can produce an instance-level semantic map layer with substantially improved 3D semantic segmentation accuracy and find complementary improvements for the fused dense semantic layer.
To the best of our knowledge, this is the first open-vocabulary semantic mapping method to perform 3D semantic fusion of dense map embeddings with instance-level map embeddings. In summary, our contributions are:

\begin{itemize}
  \item An online 3D open-vocabulary semantic voxel mapping method that simultaneously generates dense and instance-level semantic embedding layers to leverage the complementary benefits of both mapping strategies.
  \item An uncertainty-aware 3D semantic cross-layer fusion mechanism that combines embeddings of multiple, potentially overlapping instance hypotheses with those of the dense semantic map layer.
  \item A sliding window mapping approach that enables accurate instance-level semantic mapping in large scale environments by performing 3D semantic fusion between the embeddings of a dense semantic map layer, restricted to an egocentric sliding window, and a global instance-level semantic layer.
\end{itemize}

We extensively evaluate \myMethod on established 3D semantic segmentation benchmarks, comparing the outputs of our cross-layer fusion method with prior work, as well as with the unfused versions of our method's dense and instance-level layers.
We present results using two different pairs of language-aligned vision encoders, indicating that the method is not limited to a specific encoder.
Furthermore, we evaluate the sliding window mapping approach in large scale environments, reporting semantic segmentation accuracy and the memory footprint.
These experiments show that \myMethod achieves highly accurate and scalable open-set instance-level semantic mapping.
We will make the source code available \footnote{\url{https://githanonymous.github.io/FUS3DMaps/}}.

%% file: 02_related_work.tex
\section{Related Work}

Open-set semantic mapping methods commonly rely on internet-scale contrastively pre-trained vision models. To preserve the foundational generalization of these models, open-set approaches often avoid fine-tuning with labeled semantic segmentation datasets and instead use training-free strategies to spatially ground semantic embeddings~\cite{jatavallabhula2023conceptfusionopensetmultimodal3d}\cite{gu2023conceptgraphsopenvocabulary3dscene}.

\textbf{Instance-Level Semantic Maps.}
Open-Fusion \cite{yamazaki2023openfusionrealtimeopenvocabulary3d} introduces a real-time semantic TSDF voxel mapping method that uses segment-level embeddings produced by a generalist segmentation model and performs efficient Hungarian matching for segment-to-instance associations in image space.
The method assigns per-instance embedding vectors upon first observation and does not incorporate semantic updates from subsequent views, which limits the multi-view semantic consistency of the map.

ConceptGraphs~\cite{gu2023conceptgraphsopenvocabulary3dscene} and HOV-SG~\cite{werby23hovsg} generate instance-level scene-graphs. Both methods use segmentation and subsequent encoding of the corresponding image crops to generate an embedding for each instance.
HOV-SG additionally encodes the segment with the rest of the crop masked out, and the full RGB frame and averages them together with the aim of making instance semantics aware of the context the segment appears in. However, as the full frame embeddings reflect every instance in the image, averaging them together also bleeds semantics across instances.
Both of these methods are designed for offline scene-graph generation.
\cite{martins2024ovo} presents an online semantic mapping approach with a similar per-instance embedding combination strategy to \cite{werby23hovsg}. However, instead of using fixed weights as in~\cite{werby23hovsg}, they train a neural network that learns to predict weights to blend embeddings of image crops, segments with masked background, and full RGB frames.

Addressing segmentation variations from different viewpoints by maintaining multiple instance hypotheses has previously been suggested in OpenVox~\cite{deng2025openvoxrealtimeinstancelevelopenvocabulary}. However, prior to segmentation, \cite{deng2025openvoxrealtimeinstancelevelopenvocabulary} relies on an open-vocabulary object detector to propose image regions, which acts as a filter on which instance segments are processed by the pipeline.

\textbf{Dense Semantic Maps.}
ConceptFusion~\cite{jatavallabhula2023conceptfusionopensetmultimodal3d} uses segmentation and encoding of separate image crops that contain the segments. A local-to-global zero-shot embedding fusion scheme combines image crop embeddings with the embedding of the full RGB frame to generate pixel-level embeddings.
These are subsequently lifted to a dense point-cloud representation of the environment.
In \cite{Peng2023OpenScene} the authors propose to aggregate embeddings, obtained from a 2D open-vocabulary semantic segmentation model, in 3D point clouds through 3D projection and average pooling. Subsequently, they use them to supervise the training of a model that predicts semantic embeddings directly from 3D points.
However, both \cite{jatavallabhula2023conceptfusionopensetmultimodal3d} and \cite{Peng2023OpenScene} are designed as offline methods and assume upfront availability a complete sequence of RGB-D frames.

The dense semantic layer used in FUS3DMaps is similar
to~\cite{alama2025rayfrontsopensetsemanticray}. The authors introduce a novel patch-level semantic
encoder (NARADIO) which they use to construct dense 3D
semantic maps online. The encoder consists of a combination
of an agglomerative vision model backbone~\cite{heinrich2025radiov25improvedbaselinesagglomerative}, a head
that projects the features into the semantic space of a
contrastively pretrained language-aligned model~\cite{zhai2023sigmoid}, and
an additional training-free mechanism that improves the
locality of patch embeddings. \cite{alama2025radsegunleashingparametercompute} recently proposed an improved attention mechanism for the front-end vision model.
Compared to combining image crop embeddings with full image embeddings
like in \cite{werby23hovsg} and \cite{martins2024ovo}, patch-based methods can provide more
localized embeddings while also retaining some image context.
Furthermore, such methods are not affected by issues induced
by segmentation variations. However, patches may not align
with instance boundaries, and given that these methods
produce dense semantic maps with one embedding per voxel,
they scale worse with map volume than instance-based
approaches.

In \myMethod we propose to combine patch embedding based methods
with multi-hypothesis instance based methods in a principled
way, providing both accurate and scalable semantic voxel
mapping.

%% file: 03_method.tex
\begin{figure*}[t] 
    \centering
    \includegraphics[width=\textwidth]{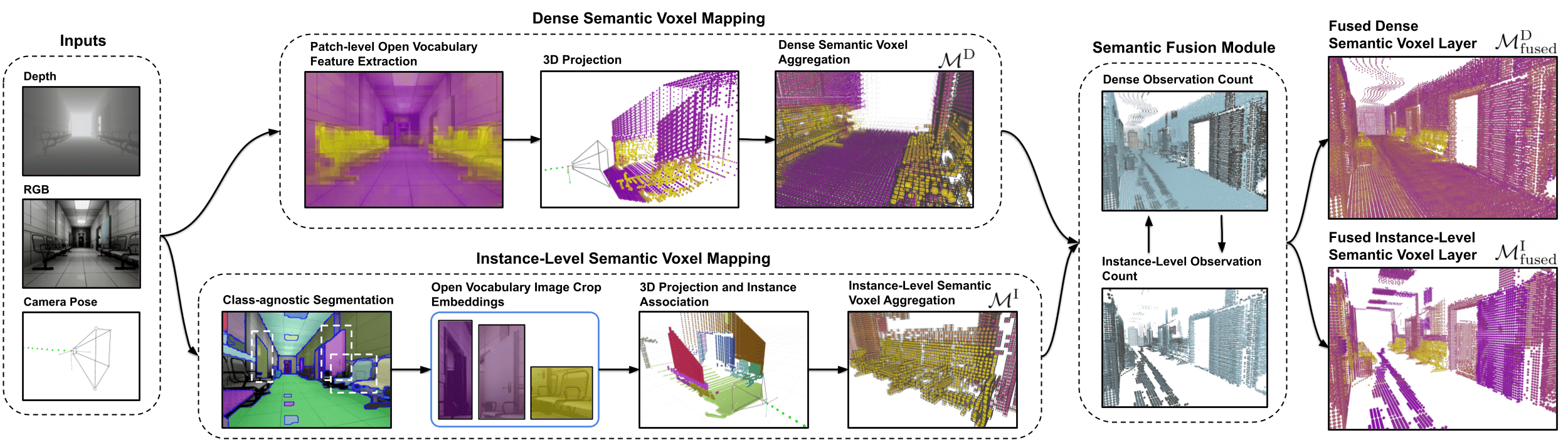}
    \vspace{-0.6cm}
    \caption{\myMethod simultaneously maintains a dense semantic layer and an instance-level semantic layer in a shared voxel map. Both mapping pipelines use encoders that produce embeddings in a shared language-aligned open-vocabulary semantic space. A cross-layer semantic fusion module allows combining embeddings of the dense and instance-level map layers to produce a fused dense semantic voxel map layer and a fused instance-level semantic voxel layer.}
    \label{fig:system framework}
    \vspace{-0.5cm}
\end{figure*}

\section{Method}
\vspace{-0.1cm}

\myMethod simultaneously maintains a dense semantic 3D voxel layer $\mathcal{M}^{\text{D}}$ with aggregated semantic patch-level embeddings and a sparser instance-level voxel layer $\mathcal{M}^{\text{I}}$, constructed using a segmentation and image crop encoding strategy. As illustrated in Fig.~\ref{fig:system framework}, our proposed cross-layer voxel-level semantic fusion method combines embeddings of $\mathcal{M}^{\text{D}}$ and $\mathcal{M}^{\text{I}}$ to produce the fused semantic map layers $\mathcal{M}^{\text{D}}_{\text{fused}}$ and $\mathcal{M}^{\text{I}}_{\text{fused}}$.
The remainder of the section is organized as follows. The dense semantic mapping pipeline is described in Section~\ref{sec:dense} and the instance-level semantic mapping pipeline in Section~\ref{sec:instance}, with
Section~\ref{sec:fuse} presenting the proposed cross-layer semantic fusion method.
Finally, Section~\ref{sec:slidingwindow} details the sliding window approach that maintains the dense layer locally and the instance-level layer globally to combine the scalability of instance-based maps with the semantic accuracy of dense semantic maps.

\vspace{-0.2cm}
\subsection{Dense Semantic Voxel Mapping}
\label{sec:dense}
\vspace{-0.1cm}

For each RGB frame $\mathcal{C}_t$, we project a subsample\footnote{The pixel sampling is detailed in Section~\ref{sec:parameter_overview}.} of its pixels to 3D using the depth frame $\mathcal{D}_t$ and the corresponding camera pose.
Additionally, we encode $\mathcal{C}_t$ with an encoder $\text{Enc}_{\text{patch}}(\cdot)$ to obtain a semantic embedding vector $\mathbf{f}^{\text{patch}} \in \mathbb{R}^d$ for each image patch in $\mathcal{C}_t$.
An embedding vector is associated with all points that were sampled and projected from the corresponding patch.
We aggregate the patch-level semantic embeddings in the dense voxel layer $\mathcal{M}^{\text{D}}$ by forming per-voxel dense embedding vectors $\mathbf{f}_{i}^{\text{D}}$ as

\begin{equation}
\label{eq:dense_agg}
\begin{aligned}
    w_{i}^\text{D}  &\gets w_{i}^\text{D} + N_i^{\text{D}}, \\
    \mathbf{f}_{i}^{\text{D}} &\gets \frac{w_{i}^\text{D} \cdot \mathbf{f}_{i}^{\text{D}} + \sum\nolimits_{j} \mathbf{f}_{j}^{\text{patch}}}{w_{i}^\text{D}},
\end{aligned}
\end{equation}
where $N_i^{\text{D}}$ is the number of points $\text{p}^{\text{D}}_{j}$ that fall within voxel $v^{\text{D}}_i$ and $\mathbf{f}_{j}^{\text{patch}}$ are the patch embeddings associated with the points $\text{p}^{\text{D}}_{j}$. The aggregated weight $w^{\text{D}}_i$ is the total number of point hits of voxel $v_i^{\text{D}}$. In the following we will refer to the accumulated point hits as the observation count of a voxel.

\subsection{Instance-Level Semantic Voxel Mapping}
\label{sec:instance}
\vspace{-0.1cm}

Alongside $\mathcal{M}^{\text{D}}$, \myMethod generates a semantically sparse instance-level semantic layer $\mathcal{M}^{\text{I}}$. 
To enhance the robustness to segmentation inaccuracy, ambiguity caused by variations in segmentation granularity, and partial occlusions, we employ a 3D multi-hypothesis voxel-mapping scheme.

\subsubsection{Instance-Level Map Structure}
$\mathcal{M}^{\text{I}}$ groups voxels into discrete instances $\mathcal{I}_{\text{id}}$. Each voxel $v^{\text{I}}_i$ stores up to $K$ instance hypotheses in the form of observation counts $w^{\text{I}}_{i,\text{id}}$ that represent evidence that the voxel $v^{\text{I}}_i$ is part of instance $\mathcal{I}_{\text{id}}$.
Furthermore, every instance $\mathcal{I}_{\text{id}}$ has an associated instance-level embedding vector $\mathbf{f}^{\text{I}}_{\text{id}} \in \mathbb{R}^d$.

\subsubsection{Segmentation and Embedding Extraction}

We utilize a class-agnostic segmentation model $\text{Seg}(\cdot)$ to generate a set of mask proposals $\mathcal{Q} = \{Q_m\}$ from an input RGB frame $\mathcal{C}_t$. We project pixels, sampled from $\mathcal{C}_t$ and contained in $Q_m$, to 3D using the depth frame $\mathcal{D}_t$. For each mask $Q_m$, this produces an instance proposal $\mathcal{P}_m$ as a set of 3D points. We filter masks with contours that have significant contact with the image border or are below a size threshold, as these often lack sufficient visual context for reliable semantic encoding.
Additionally, we extract an image crop from $\mathcal{C}_t$ that bounds $Q_{m}$ with a small padding and encode it using an image-level encoder $\text{Enc}_{\text{img}}(\cdot)$. This produces an embedding vector $\mathbf{f}^{\mathcal{P}}_\text{m} \in \mathbb{R}^d$ for each instance proposal $\mathcal{P}_m$. The image-level encoder $\text{Enc}_{\text{img}}(\cdot)$ is chosen such that $\mathbf{f}^{\mathcal{P}}_m$ resides in the same language-aligned semantic space as the embedding vectors produced by $\text{Enc}_{\text{patch}}(\cdot)$.

\subsubsection{Geometric Instance Association}
\label{sec:geom_instass}

We associate new 3D instance proposals $\mathcal{P}_m$ with existing instances through geometric overlap.
Let $\mathcal{V}_m$ denote the set of voxels occupied by $\mathcal{P}_m$. 
We compute the Intersection-over-Union (IoU) between $\mathcal{V}_m$ and the voxel sets of existing instances. If the IoU with the most overlapping instance $\mathcal{I}_{\text{id}^*}$ exceeds a threshold $\tau_{\text{iou}}$, we consider it a match.

In case of a match, we update the observation counts of the voxels $v_i \in \mathcal{V}_\text{m}$ as $w_{i,\text{id}^*}^{\text{I}} \gets w_{i,\text{id}^*}^{\text{I}} + N_i^{\text{I}}$,
where $N_i^{\text{I}}$ is the number of points $p_{j}^{\text{I}} \in \mathcal{P}_m$ that fall within voxel $v^{\text{I}}_i$.
If there is no match for $\mathcal{P}_m$, we initialize a new instance $\mathcal{I}_{\text{id}'}$ with observation counts $w_{i,\text{id}'}^{\text{I}} \gets N_i^{\text{I}}$.
If an instance is newly associated with a voxel that already maintains $K$ instance hypotheses, the hypothesis with the lowest observation count is pruned.
Maintaining per-voxel observation counts in analogous fashion to the dense mapping pipeline allows to model observational support across voxels, independent of the instance geometry, and thereby facilitates voxel-level cross-layer semantic fusion (Section \ref{sec:fuse}).

\subsubsection{Instance-Level Embedding Update}
For proposals with a match, we define $N_\mathcal{P}$ as the number of points in the instance proposal $\mathcal{P}_{m}$ and update the instance-level embeddings as

\begin{equation}
\begin{aligned}
    w_{\text{id}^*}^{\text{I}} &\gets w_{\text{id}^*}^{\text{I}} + N_\mathcal{P} \\ 
    \mathbf{f}^{\text{I}}_{\text{id}^*} &\gets \frac{w_{\text{id}^*}^{\text{I}} \cdot \mathbf{f}^{\text{I}}_{\text{id}^*} +  N_\mathcal{P} \cdot \mathbf{f}^{\mathcal{P}}_{m}}{w_{\text{id}^*}^{\text{I}}}.
\end{aligned}
\end{equation}


Consequently, embeddings that correspond to more complete views of an object receive a relatively higher weight compared to partially occluded views.
New instances are initialized with embedding $\mathbf{f}^{\text{I}}_{\text{id}'}= \mathbf{f}^{\mathcal{P}}_m$ and observation count $w^{\text{I}}_{\text{id}'} = N_\mathcal{P}$.
The voxel grids of $\mathcal{M}^{\text{I}}$ and $\mathcal{M}^{\text{D}}$ are spatially aligned. Therefore, in the following, we will refer to voxels simply as $v_i$.

\subsection{3D Cross-Layer Semantic Fusion}
\label{sec:fuse}

To combine embeddings of the dense semantic map layer $\mathcal{M}^{\text{D}}$ and the instance-level semantic map layer $\mathcal{M}^{\text{I}}$, we devise a voxel-level cross-layer semantic fusion approach.

Observation counts in our dense and instance-level semantic mapping pipelines vary across voxels depending on the scene coverage and occlusions in the processed RGB-D sequence.
The instance-level layer is additionally subject to viewpoint-dependent segmentation variations. For example, a close-up view might separately segment the keyboard and the screen of a laptop, whereas in a more distant view, the laptop may be enclosed by a single segment. This results in a non-uniform distribution of observations across voxels and instances.
Furthermore, the fidelity of the extracted semantic embeddings of the respective mapping pipeline can vary based on the vision models used.

We propose a minimal probabilistic fusion mechanism to account for these uncertainties. We model $\mathbf{f}^{\mathrm{D}}_i$ and $\mathbf{f}^{\mathrm{I}}_i$ as two independent unbiased Gaussian estimates of the true semantic embedding of a voxel $v_i$.
Each is formed by averaging $N^{\mathrm{D}}_i$ and $N^{\mathrm{I}}_i$ observations with
per-observation noise variances $R^{\mathrm{D}}$ and $R^{\mathrm{I}}$, respectively. We assume uniform $R^{\mathrm{D}}$, $R^{\mathrm{I}}$ across all observations in the respective pipeline. 
Thus, we obtain the variances of the estimates as
\begin{equation}
    \sigma_{\text{D},i}^2 = \frac{R^{\mathrm{D}}}{N_i^{\mathrm{D}}}, \qquad
    \sigma_{\text{I},i}^2 = \frac{R^{\mathrm{I}}}{N_i^{\mathrm{I}}}.
\end{equation}

Under independence, the minimum mean squared error (MMSE) estimate is the precision-weighted average
\begin{equation}
    \mathbf{f}_{\mathrm{fused},i}
    = \frac{\Pi_i^\text{D}\,\mathbf{f}^{\mathrm{D}}_i + \Pi_i^\text{I}\,\mathbf{f}^{\mathrm{I}}_i}
           {\Pi_i^\text{D} + \Pi_i^\text{I}},
    \qquad \Pi_i = 1/\sigma_i^2.
\end{equation}

Substituting $\Pi_i^\text{D} = N_i^{\mathrm{D}}/R^{\mathrm{D}}$,
$\Pi_i^\text{I} = N_i^{\mathrm{I}}/R^{\mathrm{I}}$, and
$\lambda := R^{\mathrm{D}}/R^{\mathrm{I}}$, and using $N_i^{\mathrm{D}} = w_i^{\mathrm{D}}$ and $N_i^{\mathrm{I}} = w_i^{\mathrm{I}}$, the observation counts of the voxel in the respective map layer, we obtain
\begin{equation}
    \mathbf{f}_{\mathrm{fused},i}
    = \frac{w_i^{\mathrm{D}}\cdot\mathbf{f}^{\mathrm{D}}_i
          + \lambda\,w_i^{\mathrm{I}}\cdot\mathbf{f}^{\mathrm{I}}_i}
           {w_i^{\mathrm{D}} + \lambda\,w_i^{\mathrm{I}}},
\end{equation}
where $\lambda$ encodes the relative variance of the two sources.
While the individual variances are not known, in this formulation the only parameter left is their relative variance $\lambda$, which is easy to estimate empirically by hyperparameter sweeps for any pair of vision models $\text{Enc}_{\text{patch}}(\cdot)$, $\text{Enc}_{\text{img}}(\cdot)$.

\textbf{Instance-Level Fused Semantic Layer.} 
For each instance $\mathcal{I}_{\text{id}}$ in $\mathcal{M}^\text{I}$, we first perform voxel-level semantic fusion for all voxels $v_i \in \mathcal{I}_{\text{id}}$,
\begin{equation}
    \label{eq:localinstfusion}
    \mathbf{f}^{\text{*}}_{\text{fused},i} = \frac{w_i^{\text{D}} \cdot \mathbf{f}^\text{D}_i + \lambda w_{i,\text{id}}^{\text{I}} \cdot \mathbf{f}^\text{I}_{\text{id}}}{w_i^{\text{D}} + \lambda w_{i,\text{id}}^{\text{I}}},
\end{equation}

where $w_{i,\text{id}}^{\text{I}}$ is the observation count at voxel $v_i$ corresponding to instance $\mathcal{I}_{\text{id}}$. $\mathbf{f}^\text{I}_{\text{id}}$ is the same for all voxels $v_i \in \mathcal{I}_{\text{id}}$.
To derive the instance-level fused embeddings $\mathbf{f}^{\text{I}}_{\text{fused},\text{id}}$ of $\mathcal{M}^{\text{I}}_{\text{fused}}$, we aggregate these voxel-level estimates $\mathbf{f}^{\text{*}}_{\text{fused},i}$, weighted by the total precision per voxel $\rho_i = w_i^{\text{D}} + \lambda w_{i,\text{id}}^{\text{I}}$,
\begin{equation}
\label{eq:sparse_fusion}
\mathbf{f}^{\text{I}}_{\text{fused}, \text{id}} = \frac{\sum_{v_i \in \mathcal{I}_{\text{id}}} \rho_i \cdot \mathbf{f}^{\text{*}}_{\text{fused}, i}}{\sum_{v_i \in \mathcal{I}_{\text{id}}} \rho_i}
\end{equation}

which weights voxel embeddings proportionally to the total observations they received from both sources.

\textbf{Dense Fused Semantic Layer.}
For each voxel $v_i$, the dense fused embedding $\mathbf{f}^{\text{D}}_{\text{fused},i}$ of $\mathcal{M}^{\text{D}}_{\text{fused}}$ is computed as
\begin{equation}
    \mathbf{f}^{\text{D}}_{\text{fused},i} = \frac{w_i^{\text{D}} \cdot \mathbf{f}^\text{D}_i + \lambda \hat{w}_i^{\text{I}} \cdot \hat{\mathbf{f}}^\text{I}_i}{w_i^{\text{D}} + \lambda \hat{w}_i^{\text{I}}}
\end{equation}
where $\hat{\mathbf{f}}^\text{I}_i$ is the embedding vector of the instance hypothesis with the highest observation count $\hat{w}^{\text{I}}_i$ at voxel $v_i$.

The 3D cross-layer fusion is stateless and can therefore be performed at any time to produce $\mathcal{M}^{\text{D}}_\text{fused}$ and $\mathcal{M}^{\text{I}}_\text{fused}$, for example, after receiving a text prompt or at a regular rate.

\subsection{Sliding Window Semantic Fusion}
\label{sec:slidingwindow}

Open-set semantic maps that store a high-dimensional embedding vector per voxel grow very memory intensive when scaled beyond room-sized environments.\footnote{The vision models used in our experiments produce embedding vectors of dimensions 1152 (NARADIO) and 512 (CLIP-DINOiser).}

Therefore, we propose a functionality that limits the spatial extent of the dense layer by maintaining it within a sliding window with radius $r_w$, centered at the camera position. The dual-layer mapping approach of \myMethod allows the voxels of $\mathcal{M}^{\text{D}}$ with associated dense semantic embeddings $\mathbf{f}^{\text{D}}$ to simply be pruned once they leave the window, while the instances of $\mathcal{M}^{\text{I}}$ along with the sparse raw and fused instance embeddings $\mathbf{f}^{\text{I}}$, $\mathbf{f}^{\text{I}}_{\text{fused}}$ continue to be maintained globally. Consequently, the semantic vectors scale only linearly with the number of instances $\Theta\!\left(\mathcal{\vert I \vert}\right)$ 
as in a pure instance map, rather than cubically with the resolution of the map geometry $\Theta\!\left(\frac{V}{d^3} + \mathcal{\vert I \vert}\right)
$, where $V$ is the total mapped voxel volume and $d$ is the voxel cube size.
Semantic fusion is performed at a regular rate within the sliding window as described in Section~\ref{sec:fuse}. To ensure that only instances with sufficient aggregate semantic information are updated during fusion, we maintain an evidence score for each instance $\mathcal{I}_{\text{id}}$ that is derived as
\begin{equation}
e^{\text{score}}_{\text{id}} = \sum_{v_i \in \mathcal{I}_{\text{id}}} \rho_i,
\end{equation}

where $\rho_i$ is the precision at voxel $v_i$ as used in (\ref{eq:sparse_fusion}).
Only if the new evidence score for $\mathcal{I}_{\text{id}}$, exceeds the previously stored evidence score, the fused embedding vector $\mathbf{f}^{\text{I}}_{\text{fused}, \text{id}}$ is updated along with $e^{\text{score}}_{\text{id}}$. \looseness = -1

Taken together, we refer to this mechanism as the \textit{sliding window mode}.
It enables combining
aggregated patch
embeddings of the dense layer with instance-level
embeddings, while reducing memory requirements and
limiting the computation required for semantic fusion, even if environments are very large.
A top-down view of the maps generated using an HM3DSem sequence~\cite{yadav2022habitat} is shown in Fig.~\ref{fig:hybrid}.

\subsection{Implementation Details}
\label{sec:impl}

All embeddings that are part of the voxel maps $\mathcal{M}^{\text{D}}$, $\mathcal{M}^{\text{I}}$, $\mathcal{M}^{\text{D}}_{\text{fused}}$ and $\mathcal{M}^{\text{I}}_{\text{fused}}$ share a joint semantic vision-language space.
Using a text prompt encoded with $\text{Enc}_{\text{text}}(\cdot)$, which operates in the same space, we can derive semantic similarities for all voxels using cosine similarity.

Following prior work~\cite{martins2024ovo}\cite{yamazaki2023openfusionrealtimeopenvocabulary3d}, we select a set of key-frames for processing.
To this end, we use angular and translational motion thresholding, which helps prevent the accumulation of information from redundant viewpoints, e.g. if the camera is not in motion.
Since instance-level mapping involves segmentation and image-crop encoding, which are relatively computationally expensive, we opt to process these frames less frequently by selecting higher thresholds for the instance-level mapping pipeline compared to the dense mapping pipeline. To keep the number of observations per traveled distance and rotation balanced between the two maps, we use a proportionally higher pixel sampling density for 3D projection of points in the instance-level mapping. The values used in our experiments are described in Section~\ref{sec:parameter_overview}.

%% file: 04_experimental_setup.tex
\section{Experimental Setup}

We demonstrate our proposed  cross-layer semantic fusion method for two different sets of patch-level and image-level encoders $\text{Enc}_{\text{patch}}(\cdot)$, $\text{Enc}_{\text{img}}(\cdot)$, which we detail in the following. Subsequently, we present a parameter overview and the evaluation protocol for 3D semantic segmentation.

\subsection{Vision Models}
\label{sec:vis_models}

\subsubsection{NARADIO Setup}
\label{sec:vlm2}
As the primary combination of vision-models, we employ variants of the NARADIO encoder, proposed by~\cite{alama2025rayfrontsopensetsemanticray}. NARADIO is based on an agglomerative vision foundation model, RadioV2.5~\cite{heinrich2025radiov25improvedbaselinesagglomerative}, which is trained by distilling multiple teacher models. The features are projected into the summary token (\scene{CLS}) space of SIGLIP~\cite{zhai2023sigmoid}, using a pre-trained adapter head. A spatial attention mechanism~\cite{hajimiri2025naclip} helps reinforce the spatial consistency of semantic embeddings in patch neighborhoods.
As image-level embeddings ($\text{Enc}_{\text{img}}(\cdot)$) we use the \scene{CLS} token of the SIGLIP adapter.

\subsubsection{Clip-DINOiser Setup}
\label{sec:vlm1}

In Section~\ref{sec:ablations}, we show results with a second pair of encoders.
To extract patch-level embeddings, we use CLIP-DINOiser~\cite{wysoczańska2024clipdinoiserteachingclipdino}.
The work shows that the output embeddings of MaskCLIP~\cite{zhou2022maskclip} can be refined through a guided pooling operation that is trained to replicate the patch affinity as produced by DINO~\cite{caron2021emergingpropertiesselfsupervisedvision}.
The method involves a saliency estimation step.
Rather than applying a hard threshold to the saliency estimates to extract a binary background mask as in~\cite{wysoczańska2024clipdinoiserteachingclipdino}, we use them as weights in the dense semantic mapping update (Equation~\ref{eq:dense_agg}), replacing the observation counts.
We complement the patch-level embeddings with the image-level embeddings of OpenClip~\cite{ilharco_gabriel_2021_5143773}, whose semantic space CLIP-DINOiser inherits.

\subsubsection{Segmentation Model}

For image segmentation ($\text{Seg}(\cdot)$) we use FastSAM~\cite{zhao2023fastsegment}, a class-agnostic segmentation model that adapts an efficient convolutional neural network (CNN) architecture to the \textit{Segment Anything} task introduced in~\cite{kirillov2023segany}.

\subsection{Parameter Overview}
\label{sec:parameter_overview}
In all experiments, we use a voxel side length of \qty{5}{cm} and resize depth and RGB frames to a resolution of $480 \times 640$.
For dense semantic mapping, we use tighter motion thresholds (0.08~\si{m}, 0.06~\si{rad}) than for instance-level mapping (0.14~\si{m}, 0.11~\si{rad}), but a lower pixel sampling density (5 vs. 9 pixels per image patch; patches correspond to those of $\text{Enc}_{\text{patch}}(\cdot)$).
The overlap threshold $\tau_{\text{iou}}$ for the geometrical instance association (Section.~\ref{sec:geom_instass}) is 0.2 and the maximum number of instance-hypotheses $K$ is 3 per voxel. We use a variance ratio $\lambda$ of 1.0 with NARADIO and 5.0 with the Clip-DINOiser setup.

\subsection{Dataset Experiments}

Following prior work~\cite{jatavallabhula2023conceptfusionopensetmultimodal3d}\cite{alama2025rayfrontsopensetsemanticray} we evaluate semantic 3D segmentation with a set of RGB-D sequences from the synthetic Replica dataset~\cite{replica19arxiv} (scenes: \scene{office: [0-4]}, \scene{room: [0-2]}) and real camera sequences from  Scannet~\cite{dai2017scannet} (scenes: \scene{0011}, \scene{0050}, \scene{0231}, \scene{0378}, \scene{0518}).
While Replica and ScanNet are standard data sets for semantic 3D segmentation, they consist mostly of single rooms.
To better quantify the scaling properties of the sliding window mode (Section~\ref{sec:slidingwindow}), we additionally conduct experiments with RGB-D sequences from Habitat Matterport 3D Semantic (HM3DSem)~\cite{yadav2022habitat}.
We use seven sequences (\scene{00824}, \scene{00829}, \scene{00843}, \scene{00847}, \scene{00873}, \scene{00877}, \scene{00890}), with trajectories generated by~\cite{werby23hovsg}. They include apartments and multi-story houses with, e.g., 8 rooms (\scene{00824}) and 10 rooms (\scene{00847}), respectively. We use a sliding window with a radius of \qty{6}{m} and evaluate using the 40 class labels of Matterport~\cite{yadav2022habitat}.

We use the semantic segmentation evaluation protocol from the authors of~\cite{alama2025rayfrontsopensetsemanticray}. This entails evaluating both performance over the entire set of classes of each respective dataset, as well as without the classes that can be considered to be background, such as 'floor', 'wall', 'ceiling', 'door', 'window'. Reported metrics are the mean Intersection-over-Union (mIoU), frequency-modulated mIoU (fmIoU)~\cite{jatavallabhula2023conceptfusionopensetmultimodal3d}, and mean accuracy (Acc). 
For details on the evaluation protocol, we refer to ~\cite{alama2025rayfrontsopensetsemanticray}. As in~\cite{deng2025openvoxrealtimeinstancelevelopenvocabulary}, when evaluating the instance-level semantic maps ($\mathcal{M}^{\text{I}}$, $\mathcal{M}^{\text{I}}_{\text{fused}}$), for each voxel, we use the embedding vector from the instance hypothesis with the highest observation count.

%% file: 05_results.tex
\section{Results}

We evaluate \myMethod' 3D semantic segmentation performance, focusing exclusively on training-free open-vocabulary semantic mapping approaches.

\subsection{Semantic 3D Segmentation}
\vspace{-0.1cm}

We propose grouping methods into those that produce dense semantic maps with embeddings on voxel- or point-level and methods that produce instance-level semantic maps. The methods within each of these groups share similar scaling properties in large environments. For dense semantic mapping methods, we report metrics with and without background classes, while for instance-level methods, we report them without background only.

\subsubsection{Instance-Level Semantic Maps}
\label{sec:sparse_map_eval}

We compare the instance-level output of \myMethod ($\mathcal{M}^{\text{I}}_{\text{fused}}$), using the NARADIO encoder setup, to the recent scene-graph approaches ConceptGraphs~\cite{gu2023conceptgraphsopenvocabulary3dscene} and HOV-SG\cite{werby23hovsg} and the real-time 3D semantic mapping method Open-Fusion~\cite{yamazaki2023openfusionrealtimeopenvocabulary3d}.
To generate instance proposals, OpenVox~\cite{deng2025openvoxrealtimeinstancelevelopenvocabulary} relies on an open-vocabulary object detector that is provided with a predefined list of classes\footnote{We use the author's publicly available implementation.},
thereby avoiding the challenges associated with integrating class-agnostic segmentation~\cite{zhao2023fastsegment}\cite{kirillov2023segany} into the pipeline.
We therefore include its metrics as a reference only. In addition, we report metrics for the sliding window mode of our method ($\mathcal{M}^{\text{I}}_{\text{fused}}, \textit{Sliding Window}$). As shown in Table~\ref{tab:semseg_sparse_comparison}, \myMethod substantially outperforms all instance-level semantic mapping baselines.

\begin{table}[h]
  \vspace{-0.1cm}
  \centering
  \begin{threeparttable}
      \caption{Semantic 3D Segmentation w. Instance-Level Represenations}
      \label{tab:semseg_sparse_comparison}
      \setlength{\tabcolsep}{3pt}
      \begin{tabular}{lcrrr}
      \toprule
       & \makebox[50pt]{\textbf{class-agnostic}} & & \textbf{Replica} & \textbf{ScanNet} \\
       \midrule
      \multirow{3}{*}{\makecell[l]{\textbf{OpenVox}~\cite{deng2025openvoxrealtimeinstancelevelopenvocabulary}}} 
       & & \textbf{mIoU} & 0.2427 & 0.4449 \\
       & \xmark & \textbf{fmIoU} & 0.5361 & 0.5466 \\
       & & \textbf{Acc} & 0.5995 & 0.6181 \\
       \midrule
       \midrule
       \multirow{3}{*}{\makecell[l]{\textbf{ConceptGraphs\tnote{1}}~~\cite{gu2023conceptgraphsopenvocabulary3dscene}}} 
       & & \textbf{mIoU} & 0.1163 & 0.2162  \\
       & \cmark & \textbf{fmIoU} & 0.1661 & 0.2432 \\
       & & \textbf{Acc} & 0.1980 & 0.3104 \\
       \midrule
       \multirow{3}{*}{\makecell[l]{\textbf{HOV-SG\tnote{1}}~~\cite{werby23hovsg}}} 
       & & \textbf{mIoU} & 0.1693 & 0.2679  \\
       & \cmark & \textbf{fmIoU} & 0.3145 & 0.3605 \\
       & & \textbf{Acc} & 0.3474 & 0.4417 \\
       \midrule
      \multirow{3}{*}{\makecell[l]{\textbf{Open-Fusion}~\cite{yamazaki2023openfusionrealtimeopenvocabulary3d}}}
       & & \textbf{mIoU} & 0.1848 & 0.2997 \\
       & \cmark & \textbf{fmIoU} & 0.4364 & 0.3958 \\
       & & \textbf{Acc} & 0.4887 & 0.4552 \\
      \midrule
      \multirow{3}{*}{\makecell[l]{\textbf{\myMethod} \\ \textbf{w. NARADIO} (Ours) \\($\mathcal{M}^{\text{I}}_{\text{fused}}$)}} 
       & & \textbf{mIoU} & 0.3137 & \textbf{0.4689} \\
       & \cmark & \textbf{fmIoU} & 0.5627 & \textbf{0.5231} \\
       & & \textbf{Acc} & \textbf{0.6522} & \textbf{0.6316} \\
       \midrule
      \multirow{3}{*}{\makecell[l]{\textbf{\myMethod} \\ \textbf{w. NARADIO} (Ours) \\($\mathcal{M}^{\text{I}}_{\text{fused}}$, \textit{Sliding Window})}}
       & & \textbf{mIoU} & \textbf{0.3237} & 0.4671 \\
       & \cmark & \textbf{fmIoU} & \textbf{0.5648} & 0.5161 \\
       & & \textbf{Acc} & 0.6519 & 0.6250 \\
      \bottomrule
      \end{tabular}
      {\footnotesize
      \begin{tablenotes}
        \item[1] Numbers taken from~\cite{alama2025rayfrontsopensetsemanticray}.
      \end{tablenotes}}
  \end{threeparttable}
  \vspace{-0.2cm}
\end{table}

\subsubsection{Dense Semantic Maps}

FUS3DMaps is primarily designed for instance-level semantic mapping. However, we find that cross-layer semantic fusion can also yield improvements for dense-semantic mapping.
Table~\ref{tab:semseg_comparison_merged} shows a comparison of the dense fused output ($\mathcal{M}^{\text{D}}_{\text{fused}}$) of \myMethod with the NARADIO setup with two dense semantic mapping baselines ConceptFusion~\cite{jatavallabhula2023conceptfusionopensetmultimodal3d} and Rayfronts~\cite{alama2025rayfrontsopensetsemanticray}, which is similar to our unfused dense layer.
Our method improves with respect to both baselines across the Replica and ScanNet evaluations, with (\textit{w/ bgr}) and without background (\textit{w/o bgr}) classes.

\begin{table}[h]
  \centering
  \begin{threeparttable}
      \caption{Semantic 3D Segmentation w. Dense Semantic Maps, \\[-2pt] with (w/ bgr) and without Background Classes (wo/ bgr)}
      \label{tab:semseg_comparison_merged}
      \setlength{\tabcolsep}{5pt}
      \begin{tabular}{llrrrr}
      \toprule
       & & \multicolumn{2}{c}{\textbf{Replica}} & \multicolumn{2}{c}{\textbf{ScanNet}} \\
      \cmidrule(lr){3-4} \cmidrule(lr){5-6}
       & & \textbf{w/o bgr} & \textbf{w/ bgr} & \textbf{w/o bgr} & \textbf{w/ bgr} \\
      \midrule
      
      \multirow{3}{*}{\makecell[l]{\textbf{ConceptFusion\tnote{1} } \\{\cite{jatavallabhula2023conceptfusionopensetmultimodal3d}}}} 
       & \textbf{mIoU}  & 0.2107 & 0.2038 & 0.2176 & 0.1857 \\
       & \textbf{fmIoU} & 0.3151 & 0.3575 & 0.2671 & 0.2306 \\
       & \textbf{Acc}   & 0.3565 & 0.4158 & 0.3413 & 0.2877 \\
      \midrule
      
      \multirow{3}{*}{\makecell[l]{\textbf{Rayfronts\tnote{1}}~~\cite{alama2025rayfrontsopensetsemanticray}}}
       & \textbf{mIoU}  & 0.3937 & 0.2773 & 0.4129 & 0.3229 \\
       & \textbf{fmIoU} & 0.6203 & 0.4337 & 0.4642 & 0.3904 \\
       & \textbf{Acc}   & 0.6880 & 0.5445 & 0.5676 & 0.4915 \\
      \midrule
      
      \multirow{3}{*}{\makecell[l]{\textbf{\myMethod w.} \\ \textbf{NARADIO} (Ours) \\($\mathcal{M}^{\text{D}}_{\text{fused}}$)}} 
       & \textbf{mIoU}  & \textbf{0.4117} & \textbf{0.2940} & \textbf{0.4303} & \textbf{0.3352} \\
       & \textbf{fmIoU} & \textbf{0.6416} & \textbf{0.4615} & \textbf{0.4801} & \textbf{0.3923} \\
       & \textbf{Acc}   & \textbf{0.7095} & \textbf{0.5695} & \textbf{0.5765} & \textbf{0.4959} \\
      \bottomrule
      \end{tabular}
      {\footnotesize
      \begin{tablenotes}
        \item[1] Numbers taken from~\cite{alama2025rayfrontsopensetsemanticray}.
      \end{tablenotes}}
  \end{threeparttable}
  \vspace{-0.3cm}
\end{table}

\subsubsection{Qualitative Results}

\begin{figure}[h]
    \vspace{-0.1cm}
    \centering
    \includegraphics[width=\linewidth]{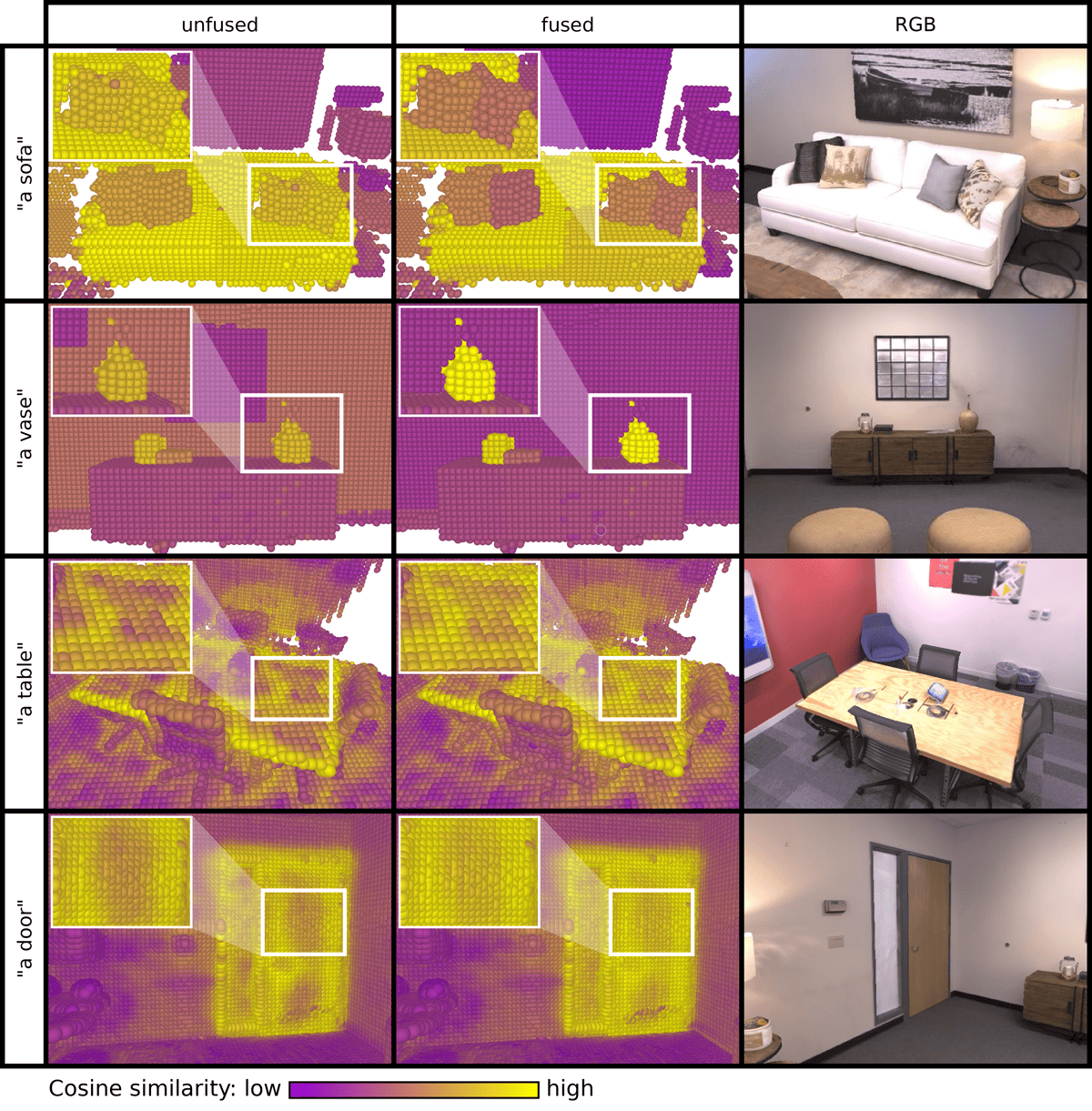}
    \vspace{-0.7cm}
    \caption{Qualitative examples of differences between the fused and unfused map layers in scene \scene{room0} of Replica. The color scheme reflects cosine similarity to the text prompt in the row label. The top two rows are examples of the instance-level layer, the bottom two rows show the dense layer.
    }
    \label{fig:retrieval}
    \vspace{-0.2cm}
\end{figure}

\begin{figure}[t]
    \centering
    \includegraphics[width=\linewidth]{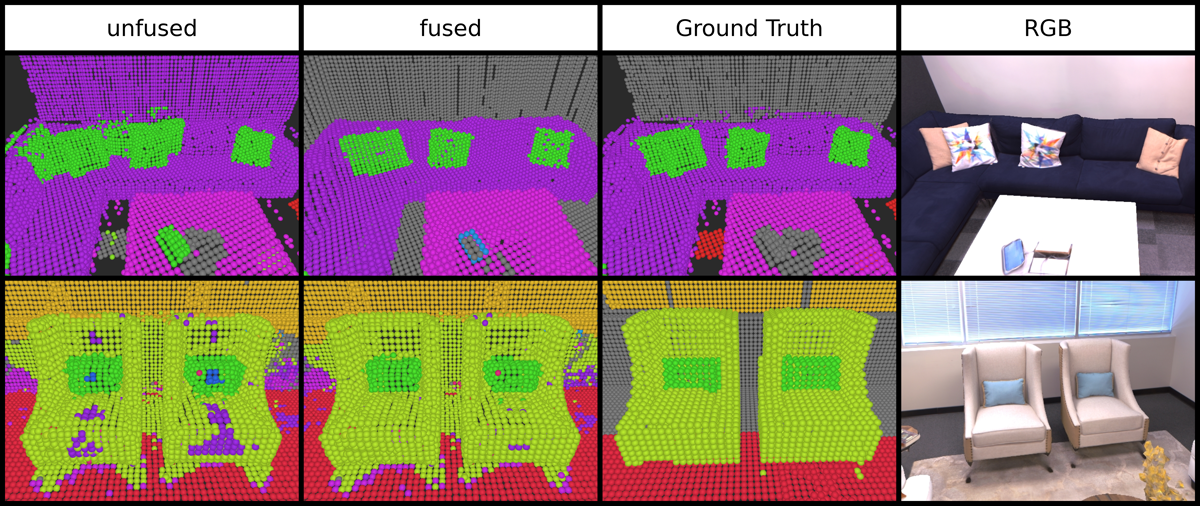}
    \vspace{-0.6cm}
    \caption{Qualitative examples of predicted class labels of the instance-level layer (top row) and dense layer (bottom row) of the voxel map.
    The colors are randomly chosen per class label and consistent between ground truth and predictions. The ground truth is visualized without background labels.}
    \label{fig:predictions_quali}
    \vspace{-0.6cm}
\end{figure}

We additionally show a set of qualitative examples on the synthetic Replica dataset with the NARADIO encoder setup. Replica was selected because the absence of geometric errors from noisy depth and drift allows visualizing the impact of the semantic fusion more clearly compared to ScanNet.
Fig.~\ref{fig:retrieval} depicts a comparison of semantic similarities with text embeddings between the fused and unfused semantic maps. The examples in the upper two rows ($\mathcal{M}^{\text{I}}$, $\mathcal{M}^{\text{I}}_{\text{fused}}$) indicate that fusion helps to distinguish pillows from cushions that are part of the sofa and helps to separate small objects, such as vases, from other objects and the background. The lower two rows ($\mathcal{M}^{\text{D}}$, $\mathcal{M}^{\text{D}}_{\text{fused}}$) show how fusion helps to reduce semantic noise within larger objects. This illustrates how fusion captures complementary strengths of instance-level and dense representations.
Similar improvements are seen for segmentation predictions in Fig.~\ref{fig:predictions_quali}.

\subsubsection{Ablations}
\label{sec:ablations}

In Tables~\ref{tab:semseg_abl_sparse} and~\ref{tab:semseg_abl_dense} we compile comparisons between the fused dense and instance-level maps, with their non-fused counterparts. We additionally show results using the alternative pair of image encoders (\textit{ClipDINO}). We find that fusion improves semantic 3D segmentation performance on nearly every metric, for both instance-level and dense maps, and across both encoder setups.

\begin{table}[h]
  \centering
  \begin{threeparttable}
      \caption{Ablation Study: Instance-Level Semantic Maps}
      \label{tab:semseg_abl_sparse}
      \begin{tabular}{llrrrr}
      \toprule
       & & \multicolumn{2}{c}{\textbf{Replica}} & \multicolumn{2}{c}{\textbf{ScanNet}} \\
      \cmidrule(lr){3-4} \cmidrule(lr){5-6}
      \textbf{} & \textbf{} & \textbf{unfused} & \textbf{fused} & \textbf{unfused} & \textbf{fused} \\
      \midrule
      \multirow{3}{*}{\makecell[l]{\textbf{\myMethod} \\ \textbf{w. NARADIO} \\ ($\mathcal{M}^{\text{I}}$/$\mathcal{M}^{\text{I}}_{\text{fused}}$)}} 
      & \textbf{mIoU} & 0.2665 & \textbf{0.3137} & 0.3808 & \textbf{0.4689} \\
      & \textbf{fmIoU} & 0.4905 & \textbf{0.5627} & 0.4227 & \textbf{0.5231} \\
      & \textbf{Acc} & 0.5687 & \textbf{0.6522} & 0.5113 & \textbf{0.6316} \\
      \midrule
      \multirow{3}{*}{\makecell[l]{\textbf{\myMethod} \\ \textbf{w. ClipDINO} \\ ($\mathcal{M}^{\text{I}}$/$\mathcal{M}^{\text{I}}_{\text{fused}}$)}}
       & \textbf{mIoU}  & \textbf{0.2499} & 0.2455 & 0.3382 & \textbf{0.3741} \\
       & \textbf{fmIoU} & 0.4203 & \textbf{0.4300} & 0.4424 & \textbf{0.4619} \\
       & \textbf{Acc}   & 0.4803 & \textbf{0.4974} & 0.5085 & \textbf{0.5350} \\
      \bottomrule
      \end{tabular}
  \end{threeparttable}
  \vspace{-0.1cm}
\end{table}

\begin{table}[h!]
  \centering
  \begin{threeparttable}
      \caption{Ablation Study: Dense Semantic Maps, \\[-2pt] with (w/ bgr) and without Background Classes (wo/ bgr)}
      \label{tab:semseg_abl_dense}
      {\setlength{\tabcolsep}{5pt}
      \begin{tabular}{llrrrr}
      \toprule
       & & \multicolumn{2}{c}{\textbf{Replica}} & \multicolumn{2}{c}{\textbf{ScanNet}} \\
      \cmidrule(lr){3-4} \cmidrule(lr){5-6}
       & & \textbf{wo/ bgr} & \textbf{w/ bgr} & \textbf{wo/ bgr} & \textbf{w/ bgr} \\
      \midrule
      \multirow{3}{*}{\makecell[l]{\textbf{\myMethod} \\ \textbf{w. NARADIO} \\ ($\mathcal{M}^{\text{D}}$)}} 
      & \textbf{mIoU} & 0.4016 & 0.2911 & 0.4045 & 0.3226 \\
      & \textbf{fmIoU} & 0.6172 & 0.4480 & 0.4635 & 0.3879 \\
      & \textbf{Acc} & 0.6848 & 0.5533 & 0.5610 & 0.4897 \\
  \cmidrule(l){2-6}
\multirow{3}{*}{\makecell[l]{\textbf{\myMethod} \\ \textbf{w. NARADIO} \\ ($\mathcal{M}^{\text{D}}_{\text{fused}}$)}} 
      & \textbf{mIoU} & \textbf{0.4117} & \textbf{0.2940} & \textbf{0.4303} & \textbf{0.3352} \\
      & \textbf{fmIoU} & \textbf{0.6416} & \textbf{0.4615} & \textbf{0.4801} & \textbf{0.3923} \\
      & \textbf{Acc} & \textbf{0.7095} & \textbf{0.5695} & \textbf{0.5765} & \textbf{0.4959} \\
      \midrule
      \multirow{3}{*}{\makecell[l]{\textbf{\myMethod} \\ \textbf{w. ClipDINO} \\ ($\mathcal{M}^{\text{D}}$)}} 
          & \textbf{mIoU}  & 0.1256 & 0.1558 & 0.2902 & 0.2769 \\
          & \textbf{fmIoU} & 0.3359 & 0.3777 & 0.3582 & 0.3450 \\
          & \textbf{Acc}   & 0.3986 & 0.4705 & 0.4370 & 0.4397 \\
      \cmidrule(l){2-6}
\multirow{3}{*}{\makecell[l]{\textbf{\myMethod} \\ \textbf{w. ClipDINO} \\ ($\mathcal{M}^{\text{D}}_{\text{fused}}$)}} 
          & \textbf{mIoU}  & \textbf{0.2657} & \textbf{0.2262} & \textbf{0.3610} & \textbf{0.3014} \\
          & \textbf{fmIoU} & \textbf{0.4684} & \textbf{0.4053} & \textbf{0.4397} & \textbf{0.3447} \\
          & \textbf{Acc}   & \textbf{0.5204} & \textbf{0.4975} & \textbf{0.5059} & \textbf{0.4306} \\
      \bottomrule
      \end{tabular}}
  \end{threeparttable}
  \vspace{-0.2cm}
\end{table}

Finally, we show parameter sweeps for the variance ratio $\lambda$ (Section~\ref{sec:fuse}) in Table~\ref{tab:lambda_abl} and report average metrics of the fused instance-level layer of all Replica and ScanNet scenes combined.

\begin{table}[h]
\vspace{-0.1cm}
\centering
\caption{Effect of $\lambda$ on Instance-Level Maps (avg. over Replica \& \\[-2pt] ScanNet)
          \textbf{Bold}: best per encoder \underline{Underline}: chosen value}
\label{tab:lambda_abl}
\vspace{-0.3cm}
\begin{tabular}{llccccc}
\toprule
& $\lambda$ & 0.1 & 0.2 & 1.0 & 5.0 & 10.0 \\
\midrule
\multirow{3}{*}{\makecell[l]{\textbf{\myMethod} \\ \textbf{w. NARADIO} \\ ($\mathcal{M}^{\text{I}}_{\text{fused}}$)}} & mIoU  & \textbf{0.391} & 0.389 & \underline{\textbf{0.391}} & 0.381 & 0.366 \\
 & fmIoU & 0.532 & 0.532 & \underline{\textbf{0.543}} & 0.517 & 0.504 \\
 & mAcc  & 0.630 & 0.630 & \underline{\textbf{0.642}} & 0.610 & 0.594 \\
\midrule
\multirow{3}{*}{\makecell[l]{\textbf{\myMethod} \\ \textbf{w. ClipDINO} \\ ($\mathcal{M}^{\text{I}}_{\text{fused}}$)}} & mIoU  & 0.243 & 0.257 & 0.306 & \underline{\textbf{0.310}} & 0.296 \\
 & fmIoU & 0.379 & 0.404 & 0.444 & \underline{\textbf{0.446}} & 0.439 \\
 & mAcc  & 0.464 & 0.490 & \textbf{0.525} & \underline{0.516} & 0.506 \\
\bottomrule
\vspace{-0.3cm}
\end{tabular}

\end{table}

\subsubsection{Framerates}

Table~\ref{tab:freq_triggered_hz} presents an overview of the average frame rates measured on the Replica and ScanNet scenes with a desktop with NVIDIA RTX 4090 and a Jetson AGX Orin 64 GB. Using motion thresholding as described in Section~\ref{sec:impl} generally leads to different frame-rates for the dense and instance-level mapping pipelines. Table~\ref{tab:freq_triggered_hz} shows the average of all processed key-frames, with the settings used in our evaluation. We report frame rates of \myMethod without fusion (\textit{no fusion}), assuming that fusion is performed upon receiving a text prompt. Furthermore, we provide frame rates using the sliding window mode, which involves semantic fusion at every fifth processed RGB-D frame (\textit{Sliding Window}), serving as an indicator of the impact of semantic fusion on the processing rate of our method.
Frame rates with NARADIO on the Jetson Orin are measured using TensorRT~\cite{NVIDIATensorRT} optimization for the backbone model.

\begin{table}[h]
\vspace{-0.1cm}
  \centering
  \begin{threeparttable}
      \caption{Average Frequency per processed Frame (Hz)}
      \label{tab:freq_triggered_hz}
      \setlength{\tabcolsep}{4pt}
      \begin{tabular}{lrrrr}
      \toprule
       & \multicolumn{2}{c}{\textbf{NARADIO}} & \multicolumn{2}{c}{\textbf{Clip-DINOiser}} \\
      \cmidrule(lr){2-3} \cmidrule(lr){4-5}
       \textbf{Setup} & \textbf{Replica} & \textbf{ScanNet}  & \textbf{Replica} & \textbf{ScanNet} \\
      \midrule
      \textbf{RTX 4090} (no fusion) & 11.85 & 11.24 & 16.50 & 15.48 \\
      \textbf{RTX 4090} (\textit{Sliding Window}) & 11.00 & 10.15 & 15.88 & 14.59 \\
      \midrule
      \textbf{Jetson} (no fusion) & 3.54 & 3.44 & 3.23 & 3.10 \\
      \textbf{Jetson} (\textit{Sliding Window}) & 3.14 & 2.94 & 3.16 & 3.00 \\
      \bottomrule
      \end{tabular}
  \end{threeparttable}
  \vspace{-0.3cm}
\end{table}

\subsection{Sliding Window Mode}

Table~\ref{tab:hm3d_comparison_avg} shows the results of experiments with seven large scale HM3DSem sequences. It compares the instance-level semantic layer ($\mathcal{M}^{\text{I}}$), the fused instance-level layer ($\mathcal{M}^{\text{I}}_{\text{fused}}$) and the fused instance-level layer produced by the sliding window mode ($\mathcal{M}^{\text{I}}_{\text{fused}}$, \textit{Sliding Window}). Additionally, we include an ablation without the evidence scoring mechanism of the sliding window mode ($\mathcal{M}^{\text{I}}_{\text{fused}}$, \textit{Sliding Window}, \textit{no} $e_{\text{id}}^{\text{score}}$), described in Section~\ref{sec:slidingwindow}. We add results for the strongest of the instance-level baselines~\cite{yamazaki2023openfusionrealtimeopenvocabulary3d} as a reference.
The results show that the sliding window mode does not significantly reduce the instance-level 3D semantic segmentation performance, compared to the full dual-layer mapping.

\begin{table}[h]
\centering
\begin{threeparttable}
\caption{Semantic 3D Segmentation Performance on \\[-2pt] HM3DSem sequences (avg. over 7 scenes)}
\label{tab:hm3d_comparison_avg}
\begin{tabular}{llr}
\toprule
\multirow{3}{*}{\makecell[l]{\textbf{Open-Fusion}~\cite{yamazaki2023openfusionrealtimeopenvocabulary3d}}}
& \textbf{mIoU}  & 0.2587 \\
& \textbf{fmIoU} & 0.3031 \\
& \textbf{Acc} & 0.3797 \\
\midrule
\multirow{3}{*}{\makecell[l]{\textbf{\myMethod w.} \\ \textbf{NARADIO} \\ ($\mathcal{M}^{\text{I}}$)}}
& \textbf{mIoU}  & 0.3553 \\
& \textbf{fmIoU} & 0.4336 \\
& \textbf{Acc} & 0.5538 \\
\midrule
\multirow{3}{*}{\makecell[l]{\textbf{\myMethod w.} \\ \textbf{NARADIO} \\ ($\mathcal{M}^{\text{I}}_{\text{fused}}$)}}
& \textbf{mIoU}  & \textbf{0.4062} \\
& \textbf{fmIoU} & \textbf{0.4708} \\
& \textbf{Acc} & \textbf{0.5887} \\
\midrule
\multirow{3}{*}{\makecell[l]{\textbf{\myMethod w.} \\ \textbf{NARADIO} \\ ($\mathcal{M}^{\text{I}}_{\text{fused}}$, \textit{Sliding Window})}}
& \textbf{mIoU}  & 0.4022 \\
& \textbf{fmIoU} & 0.4622 \\
& \textbf{Acc} & 0.5778 \\
\midrule
\multirow{3}{*}{\makecell[l]{\textbf{\myMethod w.} \\ \textbf{NARADIO} \\ ($\mathcal{M}^{\text{I}}_{\text{fused}}$, \textit{Sliding Window}, no $e_{\text{id}}^{\text{score}}$)}}
& \textbf{mIoU}  & 0.3530 \\
& \textbf{fmIoU} & 0.4135 \\
& \textbf{Acc} & 0.5278 \\
\bottomrule
\end{tabular}
\end{threeparttable}
\vspace{-0.5cm}
\end{table}

Table~\ref{tab:memory_final} shows a comparison of the memory required to maintain the dense semantic layer $\mathcal{M}^{\text{D}}$ with 3D voxel locations and dense semantic embeddings. It is compared with the memory footprint of the \myMethod sliding window mode (\textit{SW}), using a radius of $6m$.
These numbers show the total required memory, including voxel locations, observation counts, raw and fused instance embeddings, and dense embeddings within the sliding window.
The data are measured after processing the last frame of the respective dataset sequence.
Additionally, we show the memory expense of a sequence (\scene{Hospital diff P1004}) from TartanGround~\cite{patel2025tartanground}, which covers three levels of a hospital building and a sequence that we collected using a LiDAR-RGB-D sensing module~\cite{ManifoldOdin1} covering two floors of a university building. In all sequences, we use a voxel side-length of 5cm.

\begin{table}[h]
\vspace{-0.1cm}
  \centering
  \begin{threeparttable}
      \caption{Memory comparison of Semantic Maps (in MiB)}
      \label{tab:memory_final}
      
      \setlength{\tabcolsep}{8pt} 
      
      \begin{tabular}{l rr | rr}
      \toprule
       & \multicolumn{2}{c}{\textbf{NARADIO}} & \multicolumn{2}{c}{\textbf{ClipDINO}} \\
      \cmidrule(lr){2-3} \cmidrule(lr){4-5}
       \textbf{Scene} & \textbf{$\mathcal{M}^{\text{D}}$} & \textit{SW} & \textbf{$\mathcal{M}^{\text{D}}$} & \textit{SW} \\
      \midrule
      \textbf{HM3DSem 00824} & 1243 & 355 & 623 & 180 \\
      \textbf{HM3DSem 00829} & 899  & 268 & 451 & 137 \\
      \textbf{HM3DSem 00843} & 1145 & 254 & 574 & 130 \\
      \textbf{HM3DSem 00847} & 1201 & 546 & 603 & 276 \\
      \textbf{HM3DSem 00873} & 1194 & 311 & 599 & 158 \\
      \textbf{HM3DSem 00877} & 1162 & 323 & 583 & 164 \\
      \textbf{HM3DSem 00890} & 1473 & 231 & 739 & 118 \\
      \textbf{Tartanground Hospital} & 5970 & 354 & 2994 & 185 \\
      \textbf{Odin1 University} & 5862 & 1054 & 2940 & 534 \\
      \bottomrule
      \end{tabular}
  \end{threeparttable}
  \vspace{-0.2cm}
\end{table}

The results show that, at equal voxel resolution, the memory required for our dual-layer sliding window mode is substantially lower and scales better to larger environments compared to the memory required for a dense semantic map.

%% file: 06_conclusion.tex
\section{Conclusions}

We present an online open-vocabulary semantic mapping method that fuses embeddings of instance hypotheses with those of a dense semantic voxel layer. The proposed sliding window semantic mapping approach enables strong open-vocabulary semantic mapping by enriching instance-level embeddings with embeddings of the dense map layer. Furthermore, it scales substantially better with environment size than purely dense semantic maps.